\documentclass[preprint,12pt]{elsarticle}

\usepackage{amssymb}
\usepackage{amsmath}
\usepackage{multirow}
\usepackage{algorithm}
\usepackage{algpseudocode}
\usepackage{hyperref}

\journal{Arxiv}

\begin{document}

\begin{frontmatter}



\title{Multi-scale time-stepping of Partial Differential Equations with Transformers}


\author[aff1]{AmirPouya Hemmasian}
\author[aff1,aff2]{Amir Barati Farimani}

\affiliation[aff1]{organization={Department of Mechanical Engineering, Carnegie Mellon University},
            addressline={5000 Forbes Avenue}, 
            city={Pittsburgh},
            postcode={15213}, 
            state={PA},
            country={USA}}

\affiliation[aff2]{organization={Machine Learning Department, Carnegie Mellon University},
            addressline={5000 Forbes Avenue}, 
            city={Pittsburgh},
            postcode={15213}, 
            state={PA},
            country={USA}}

\begin{abstract}
Developing fast surrogates for Partial Differential Equations (PDEs) will accelerate design and optimization in almost all scientific and engineering applications.  Neural networks have been receiving ever-increasing attention and demonstrated remarkable success in computational modeling of PDEs, however; their prediction accuracy is not at the level of full deployment. In this work, we utilize the transformer architecture, the backbone of numerous state-of-the-art AI models, to learn the dynamics of physical systems as the mixing of spatial patterns learned by a convolutional autoencoder. Moreover, we incorporate the idea of multi-scale hierarchical time-stepping to increase the prediction speed and decrease accumulated error over time. Our model achieves similar or better results in predicting the time-evolution of Navier-Stokes equations compared to the powerful Fourier Neural Operator (FNO) and two transformer-based neural operators OFormer and Galerkin Transformer. The code will be available on \href{https://github.com/BaratiLab/MST_PDE}{https://github.com/BaratiLab/MST\_PDE}.

\end{abstract}




\end{frontmatter}


\section{Introduction}

Partial Differential Equations (PDEs) govern the dynamics of continuous physical systems in science and engineering. To solve them numerically, functions are represented as finite sets of numbers, typically through methods like finite difference/element/volume or spectral methods. While these methods offer stability and accuracy, they can be computationally intensive, especially for complex systems. In recent years, Machine Learning and Deep Learning have emerged as alternative approaches for scientific computation and PDE modeling due to their remarkable achievements in various fields \cite{brunton2019data, frank2020machine, cuomo2022scientific}.

The art of deep learning is to design or utilize the network architecture best suited for the data or the task at hand. Convolutional Neural Networks (CNNs) have been utilized for modal analysis and model reduction of physical systems, as well as PDE solving \cite{Bhatnagar-CNN-PDE-2019, lee2020model, fukami2020convolutional, gao2021phygeonet}. While CNNs have been shown to have adequate sample efficiency and accuracy, they are limited to data represented on an equispaced mesh. Recurrent Neural Networks (RNNs) are another class of networks that are usually used to model dynamic systems and time-dependent differential equations \cite{mohan2018deep, hasegawa2020cnn, maulik2021reduced}. Despite the success of RNNs in many applications, there are challenges with training that limit their utilization in some problems. 

The key component of this work is a novel architecture that has been achieving remarkable success in numerous applications, and that is the transformer architecture \cite{Attention-NIPS-2017, BERT, AlphaFold2, Visual-Transformer-ICLR-2021}. The transformer was first introduced as a mechanism for learning from a global context without suffering from the shortcomings of sequential models like RNNs. We will briefly review the common data-driven frameworks in PDE modeling, followed by a more focused elaboration on the applications of the transformer architecture in the context of PDE modeling.

For the sake of generality and consistency throughout the paper, we assume a PDE of the form shown in equation \ref{eq:pde} where $\textbf{x}=[x_1, x_2, ..., x_d]^T\in\mathbb{R}^d$ represents a $d$-dimensional spatial domain and $t\in\mathbb{R}$ represents time.
\begin{equation}
u_t=N(\textbf{x}, u, u_{x_1}, u_{x_2}, ..., u_{x_1x_1}, u_{x_1x_2}, u_{x_2x_2}, ...)
\label{eq:pde}
\end{equation}

\textbf{Reduced-order models.}
The Reduced-Order Modeling (ROM) framework is based on the separation of variables space and time \cite{brunton2019data} and is a classic method to obtain low-cost surrogate models of PDEs, especially fluid flows. A ROM assumes a solution form shown in equation \ref{eq:rom} where $\boldsymbol{\cdot}$ is the inner product and $a,\psi\in\mathbb{R}^K$ are the temporal and spatial components respectively. The spatial component $\psi$ consists of basis functions $\psi_k$, also known as spatial modes, that are chosen to preserve the most spatial information and variance of the system and manifest as dominant patterns.  Eventually, the temporal component $a(t)$ is obtained using analytical, numerical, or data-driven algorithms. In data-driven ROMs, sample solution functions are represented in a discretized temporal and spatial domain. Proper Orthogonal Decomposition is a classic algorithm to obtain the spatial component from the eigenvectors of the data matrix \cite{berkooz1993proper, anttonen2003pod}. Different machine learning algorithms and deep learning architectures can be utilized to model either the spatial or the temporal component of the ROM \cite{hasegawa2020cnn, murata2020nonlinear, lee2020model, hesthaven2018non, maulik2021reduced, mohan2018deep, brunton2020machine}.
\begin{equation}
u(\textbf{x},t)=a(t)\boldsymbol{\cdot}\psi(\textbf{x})=\sum_{k=1}^Ka_k(t)\psi_k(\textbf{x})
\label{eq:rom}
\end{equation}

\textbf{Latent space time-stepping.}
This framework is a generalized version of ROM where the spatial information is not simply compressed into a few modes or bases, but possibly a richer representation learned by a neural network \cite{wiewel2019latent, lee2021deep, khoo2021solving}. It usually consists of three main components namely the encoder, decoder, and dynamical model, denoted as $P, Q, D^{\Delta t}$ respectively such that $z_t=P(s_t)$, $Q(z_t)=\tilde{s}_t$ and $z_{t+\Delta t}=D^{\Delta t}(z_t)$. Here, $s_t$, $\hat{s}_t$, and $z_t$ are the state, reconstructed state, and latent/encoded state of the system at time $t$ respectively, and $z_t$ is a generalization of $a(t)$ in ROMs.

\textbf{Direct time-stepping.}
If $P$ and $Q$ from the previous framework are identity functions, only a dynamical model is to be learned to simulate the time evolution of the system in its original representation. Since the state representation is maintained, the choice of architecture suited for the problem at hand may be more transparent, i.e. using CNNs for regular meshes and GNNs for irregular meshes \cite{gupta2022multispatiotemporalscale, janny2023eagle, li2022fgn, PDE-ROM-2021, pfaff2021learning, sanchezgonzalez2020learning, belbute2020combining}. These methods usually can achieve high accuracy and learn the fine changes and patterns as well due to their working with the original representation rather than a latent or reduced one, but they come with a high computational cost for both training and testing compared to the previous frameworks.

\textbf{Solution function approximation.}
Another recently popular approach is to approximate the solution function by a parameterized function of a certain form like a neural network such that $u(\textbf{x}, t)\simeq f_\phi (\textbf{x}, t)$. The model parameters can then be obtained by minimizing the approximation error consisting of data-driven loss and physics-informed loss, giving them the name physics-informed neural networks (PINN) \cite{raissi2019physics, physics-informed-ml-review, cai2021physics, pang2019fpinns}. These models are completely mesh-agnostic and can even approximate the solution function solely from the physics-informed loss without any sampled simulation data \cite{sun2020surrogate}. However, a new model has to be trained for each instance of the PDE with new parameters and initial and boundary conditions, and the calculation of high-order derivatives in high-dimensional domains can be prohibitive.

\textbf{Neural operators.}
While previous frameworks treat the discretized representation of functions as members of Euclidean spaces, neural operators conserve their identity and properties as functions, enabling them to work with arbitrary discretizations. Based on the universal operator approximation theorem for neural networks \cite{Universal-apprx-operator-IEEE-1995}, Deep Operator Networks (DeepONets) realized the practical application of neural networks in operator learning \cite{DeepONet-Nature-2021}. Another class of neural operators uses an analogous architecture to fully connected networks by approximating nonlinear operators with a composite of several layers of kernel integral operators and nonlinear activation functions. The kernel can be approximated with a neural network \cite{li2020gno}, the column space of attention weights in a transformer \cite{cao2021choose, li2022transformer}, wavelet bases \cite{gupta2021multiwaveletbased}, or Fourier bases \cite{li2020fourier, wen2022ufno, tran2023factfno}.

\textbf{Transformer and PDE simulation.}
Following the groundbreaking success of the transformer model since its introduction \cite{AlphaFold2, Attention-NIPS-2017, BERT, Visual-Transformer-ICLR-2021, carion2020detr}, researchers have utilized its architecture and the attention mechanism to model and predict physical systems \cite{geneva2022transformers} in frameworks such as ROM \cite{solera2023beta, hemmasian2023reduced} and neural operators \cite{cao2021choose, guo2022transformer, hao2023gnot, li2023transformer, ovadia2023vito, kissas2022learning, liu2022ht, nguyen2022fourierformer, li2023scalable, han2022predicting}. As pointed out by the aforementioned works, the attention mechanism can be viewed as a discretized approximation of a kernel integral operator, making it a suitable tool in operator learning. However, the computational cost can be hindering for high-resolution meshes. The problem of long input sequences and the high computational cost of the attention mechanism has been the focus of many works and is usually mitigated by modifying the attention mechanism to have linear or quasi-linear cost in terms of the input length \cite{rae2019compressive, zaheer2021bigbird, beltagy2020longformer, Kitaev2020Reformer, shen2020efficient}.

In the context of solving PDEs, Cao \cite{cao2021choose} introduced the Galerkin Transformer which omits the softmax function in the attention and changes the order of calculations to have a linear cost in the input length, but the cost is still prohibitive in high-dimensional and high-resolution problems. Li et al. \cite{li2023scalable} applied the attention separately across different axes, reducing the total cost. Unlike such works that focus on modifying the algorithm and mechanism of the attention, ROMER \cite{hemmasian2023reduced} approaches this issue from a different perspective, investigating how compressing the data representation itself can be a path to an efficient and accurate model based on attention. This work provides this framework with essential enhancements that enable it to achieve remarkable performance, similar or superior to state-of-the-art models like neural operators. The introduced improvements consist of finite backpropagation in time, multi-scale time-stepping, and a more intuitive positional encoding for the transformer architecture, which will be explained in detail.

\section{Methodology}
This section explains the architecture of our model step by step. First, the convolutional autoencoder (CAE) learns a coarse-meshed feature representation of the system called the encoding. Next, the attention mechanism which is the backbone of the transformer architecture is introduced and its connection to the kernel integral operator is elucidated. After that, the multi-head self-attention layer and the transformer architecture are presented which are used to learn the dynamics in the encoding space. Finally, details and strategies for the training process are provided.

\subsection{Convolutional autoencoder}

We leverage the locality and multiscale nature of the spatial patterns commonly observed in physical systems like fluid flows and reduce the spatial dimension of the state representation using a CAE. A fully convolutional network learns a feature representation that conserves the order of patterns while compressing the spatial dimensions. Since the attention mechanism has a quadratic cost in input length, this greatly reduces the cost for the following components of the model based on attention.

Another important motivation to put the CAE at the stem of our model is to process information and learn features in the spatial dimensions first, rather than doing so in the temporal dimension. We believe this to be more intuitive considering the nature of a typical time-dependent PDE of the form in equation \ref{eq:pde}. This makes it more similar to classic numerical algorithms since they usually estimate spatial gradients and the right hand of this equation using discretized approximations first, then use them to predict the values at the next time-step. Likewise, our model first learns spatial features and patterns from a single time frame and models the dynamics as exchanges among them. 

The CAE consists of the encoder and the decoder, which can be viewed analogously to the lifting and projecting layer ($P$ and $Q$) at the beginning and end of neural operators like FNO \cite{FNO-ICLR-2021}. While in neural operators these layers usually operate on a concatenation of consecutive snapshots and learn temporal features only, our model works with a single snapshot in time and learns spatial features. Without the loss of generality, we present the data pipeline for a 2D spatial domain and denote the state and the encoded state as $s\in \mathbb{R}^{N_x\times N_y}$ and $z\in\mathbb{R}^{n_x\times n_y\times d_f}$ respectively. The dynamics are then to be learned in the $z$-space by the attention mechanism and the transformer model, which are explained next. 

\begin{figure}
\centering
\includegraphics[width=1\linewidth]{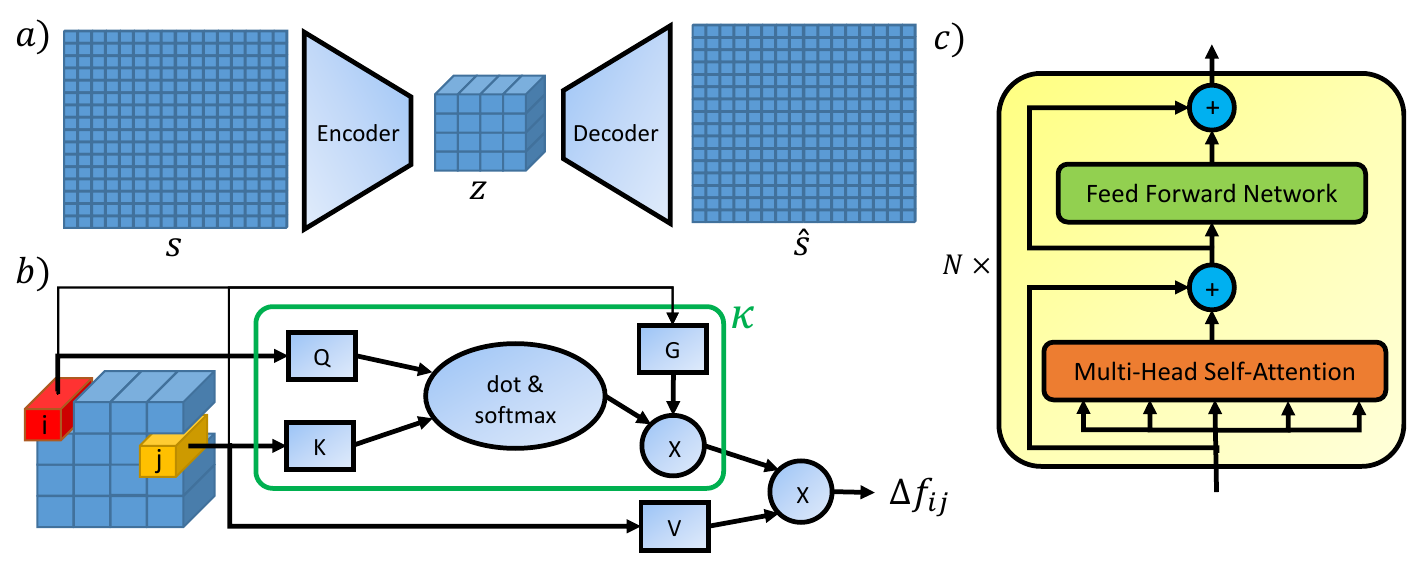}
  \caption{a) The convolutional autoencoder at the stem of our model. b)The attention mechanism of our model with disentangled incorporation of positional information and feature values. c) The transformer model consisting of $N$ transformer layers. The feed-forward network is implemented using 1x1 convolution layers to apply an identical fully connected network to all elements.}
\label{fig:model}
\end{figure}

A simple visualization of the CAE and its functionality is provided in Figure \ref{fig:model}a. The detailed architecture is similar to  \cite{hemmasian2023reduced}, each consisting of 4 encoder or decoder blocks respectively. An encoder block is composed of a convolution layer, an average pooling layer, and a nonlinear activation function. A decoder block is composed of a linear upsampling layer, a convolution layer, and a nonlinear activation function. Leaky ReLU is the choice for the activation function since it does not have a saturation region that causes gradient vanishing, excluding the final block of both models which do not have an activation function. All convolution layers use kernel size 3, and both pooling and upsampling layers use a scale of 2.

\subsection{The attention mechanism}

In order to pass the encoded state $z$ to the attention mechanism, it is reshaped as a set of $n=n_xn_y$ feature vectors denoted as $f_i\in \mathbb{R}^{d_f}$ where $i=0, 1, 2, ..., n-1$. We also define a corresponding set of vectors containing positional information about the location of $f_i$ vectors as $p_i\in \mathbb{R}^{d_p}$. A simple choice for $p$ would be $p=[x,y]\in \mathbb{R}^2$ containing the index or value in $x$ and $y$ axis. A general formulation of the attention mechanism is presented in equation \ref{eq:attn_formula}.
\begin{equation}
\Delta f_i := \sum_{j=0}^{n-1}\kappa(p_i, p_j, f_i, f_j)V(f_j), \hspace{1cm} \forall i\in \{0,1,...,n-1\}
\label{eq:attn_formula}
\end{equation}
Here, $\kappa:\mathbb{R}^{2(d_p+d_f)}\rightarrow\mathbb{R}$ is the attention weight function, $V:\mathbb{R}^{d_f}\rightarrow\mathbb{R}^{d_f}$ is called the value function, and $\Delta f_i$ is the incremental change of $f_i$. This formula can be viewed as a discretized version of a kernel integral operator $K_\phi$ used in neural operators like FNO \cite{FNO-ICLR-2021}, as shown in equation \ref{eq:ki_operator}.
\begin{equation}
(K_\phi(a)v)(x) := \int_{D}  \kappa_{\phi}(x,\tilde{x},a(x),a(\tilde{x}))v(\tilde{x})d\tilde{x}, \hspace{1cm} \forall x \in D
\label{eq:ki_operator}
\end{equation}
Here, $\kappa_\phi:\mathbb{R}^{2(d_x+d_a)} \rightarrow \mathbb{R}^{d_v\times d_v}$ is a parameterized kernel function and $v$ is the input to the operator. There are interesting correspondences between the two formulations; $x, \tilde{x}, v(\tilde{x})$ are analogous to $p_i, p_j, f_j$ in equation \ref{eq:attn_formula} respectively. The spatial variable is denoted as $x$ in equation \ref{eq:ki_operator} which can be of any number of dimensions, while we denote it as $p$ for position to avoid confusion with the first spatial dimension. In equation \ref{eq:ki_operator}, $a$ is the model's input which can be initial or boundary conditions or the parameters of the PDE, and it does not have an exact counterpart in equation \ref{eq:attn_formula}. The initial condition is, however, of the same shape and nature as the output of each layer and can be treated as the input to the operator itself.

The original attention \cite{Attention-NIPS-2017} and ROMER \cite{hemmasian2023reduced} assume $\kappa$ of the form shown in equation \ref{eq:attn_classic}, where $Q, K$ are linear functions named query and key, and $\boldsymbol{\cdot}$ is the dot product. The value function $V$ is also linear and takes $f+p$ as input instead of $f$. Basically, the positional information is incorporated into the model by being added to the input vectors at the beginning. This can be problematic since it limits the shape of the positional feature to be the same as the main features and also deviates the values of the feature vectors for the upcoming calculations later in the model. 
\begin{equation}
\kappa(p_i, p_j, f_i, f_j) = Q(f_i+p_i)\boldsymbol{\cdot}K(f_j+p_j)
\label{eq:attn_classic}
\end{equation}
our model assumes a different form shown in equation \ref{eq:attn_us} where $G:\mathbb{R}^{2d_p}\rightarrow \mathbb{R}$ is the positional encoder modeled by a fully connected network. This formulation disentangles the effect of positional and feature information and conserves the actual value of the input vectors for further calculations. A graphic representation of this attention mechanism is illustrated in Figure \ref{fig:model}b.
\begin{equation}
\kappa(p_i, p_j, f_i, f_j) = G(p_i, p_j) (Q(f_i)\boldsymbol{\cdot}K(f_j))
\label{eq:attn_us}
\end{equation}

A desirable and intuitive property for the positional encoder is to be a function of the relative location of $p_j$ to $p_i$, leading us to choose the formulation shown in equation \ref{eq:rel_pos}. This also makes the overall model conserve the property of translation invariance, an important and helpful inductive bias for physical systems and PDE solution operators. More arguments can also be included that conserve this property based on the design of $p$.
\begin{equation}
G(p_i, p_j)=G(p_j-p_i, |p_i-p_j|)
\label{eq:rel_pos}
\end{equation}

\subsection{Transformer}
The architecture chosen to model the dynamics in the encoding space is based on the classic transformer architecture \cite{Attention-NIPS-2017} excluding normalization layers. The architecture of the transformer is shown in Figure \ref{fig:model}c. The multi-head attention is simply a set of independent attention mechanisms that have their output concatenated together, each called an attention head. Each head learns distinct $Q, K, V, G$ which helps the expressive ability of the model as opposed to learning one positional encoder $G$ which may limit the ability to capture a diverse range of dynamics in the spatial domain. For example, one head can focus on close-range dynamics and one can learn longer-range dynamics, where $G$ assigns bigger weights to closer and further values respectively. The attention mechanism models the interaction of the feature vectors in space, the fully connected layers model the dynamics of individual elements over time, and the residual connection is meant to model time-stepping and incremental change of the system in each time-step.

\subsection{Assembly and training}

The whole model consists of an autoencoder and several transformers, each meant to model the dynamics at a specific time scale. This was inspired by the work of Liu et al. \cite{liu2022hierarchical} which took a similar approach but with low-order non-linear models and proved the concept using fully connected neural networks. We denote the encoder and decoder as $P, Q$ and the model that propagates the encoded state to the future for $\Delta t$ time-steps as $D^{\Delta t}$. This work uses up to four different dynamical models with $\Delta t=\{1,2,4,8\}$. 

The loss function used to train all models is the normalized Root Mean Squared Error (nRMSE) defined in equation \ref{eq:nRMSE}, where $\tilde{x}$ and $x$ represent the model's output and the ground truth respectively. This loss provides a measure of error not affected by the magnitude or resolution of data, making it a popular choice to evaluate PDE solvers \cite{FNO-ICLR-2021}. We use the Adam optimization algorithm \cite{Adam} to minimize the loss with the initial learning rate 0.001 and a learning rate scheduler that reduces the learning rate by a factor of 0.2 where the learning curve reaches a plateau, with a patience factor of 5.
\begin{equation}
nRMSE(\tilde{x}, x)=\frac{\big\|\tilde{x}-x\big\|_2}{\bigl\|x\bigr\|_2}
\label{eq:nRMSE}
\end{equation}

First, the autoencoder is trained to extract spatial features and dominant patterns. The loss function for the autoencoder is defined by setting $x=s$ and $\tilde{x}=Q(P(s))$ in equation \ref{eq:nRMSE}. After training, the encoder and decoder are used to move to and from the encoding space, in which the dynamics are to be learned by the transformer models.

The calculation of the loss function for the dynamical models is shown in algorithm \ref{alg:dyn_loss}. An important training strategy of this work is using rollout in training, also known as backpropagation through time. Models in similar applications usually use a full rollout ($R=T-1$) which prevents parallelized training across time hence increasing training time, and introduces a risk of gradient explosion or vanishing as well. Unlike the original applications of many recurrent models like natural language processing (NLP), a physical system has the Markov property, meaning that its future is dependent only on the present and not the past provided a sufficient state representation. Therefore, a full rollout may cost too much compared to its advantage. By using a finite-time rollout, our model can benefit from the advantages while still being able to parallelize the training over time with much less risk of gradient explosion or vanishing.

\begin{algorithm}
\caption{Calculation of the loss function of $D^{\Delta t}$ for a sample trajectory of length $T$, using a trained encoder $P$ and training rollout $R$} 
\begin{algorithmic}[1]
\For{$t=0,1,\ldots,T-R\Delta t$}
    \State $Loss=0$
    \State $z_{t}=P(s_t)$
    \State $\tilde{z}_{t}=z_{t}$
    \For{$r=1,2,\ldots,R$}
        \State $z_{t+r\Delta t}=P\bigl(s_{t+r\Delta t}\bigr)$
        \State $\tilde{z}_{t+r\Delta t}=D^{\Delta t}\bigl(\tilde{z}_{t+(r-1)\Delta t}\bigr)$
        \State $Loss\leftarrow Loss + nRMSE\bigl(\tilde{z}_{t+r\Delta t}, z_{t+r\Delta t}\bigr)$
    \EndFor
    \State $Loss \leftarrow Loss\div R$
\EndFor
\end{algorithmic} 
\label{alg:dyn_loss}
\end{algorithm}

We also use transfer learning to ease the training for the dynamical models. The dynamics for a large time-step are expected to be more complex and difficult than for small time-steps. Therefore, we use the trained dynamical models for better initialization of the next models to train for larger time-steps. For example, $D^2$ is initialized by the trained weights of $D^1$, $D^4$ by the trained $D^2$, and $D^8$ by $D^4$.

\section{Experiments and discussion}
In order to assess the capability of our model, we train it on two challenging datasets used to evaluate powerful and novel PDE solvers like neural operators. We choose 2D incompressible Navier-Stokes (NS) with three different Reynolds numbers and a low-Reynolds 2D Kolmogorov Flow (KF), both provided publicly by Li et al. in \cite{li2020fourier}. Each of the NS datasets consists of 1000 training samples and 200 test samples, and the KF dataset consists of 160 training samples and 40 test samples. The spatial resolution in all cases is $64\times 64$. 

We start with the NS datasets and compare the performance of our model with TF-Net, the powerful FNO, and two transformer-based neural operators. The transformer used for all dynamical models on NS datasets has 4 layers and 8 heads, and each component of the models is trained for 100 epochs with a batch size of 64. Looking at the results in Table \ref{table:results}, it is observed that our model performs almost as well as FNO on NS1, and outperforms all models on NS2 and NS3 with an impressive margin considering the turbulence and relatively small dataset size. It is worth mentioning that all competing models use a full rollout in training and take as input the state at 10 consecutive time-steps, while our model uses a finite rollout (sometimes 2 or 4 is enough) and takes as input the state at a single time-step.

\begin{table}
\centering
\caption{Benchmarks on 2D Navier-Stokes data provided by Li et al. \cite{FNO-ICLR-2021}. Results reported by the papers FNO \cite{FNO-ICLR-2021} and OFormer \cite{li2023transformer}.}
\begin{tabular}{|c|c|c|c|c|c|} \hline
data & NS1 & NS2 & NS3 & \#param(M) \\
 $\nu$ & $1e-3$ & $1e-4$ & $1e-5$ & ---\\
$T(T_{pred})$ & 50 (40) & 30 (20) & 20 (10) & ---\\ \hline
TF-Net \cite{TF-NET-SIGKDD-2020} & 2.25\% & 22.53\%  & 22.68\% & 7.45\\
FNO-3D \cite{FNO-ICLR-2021} & \textbf{0.86\%} & 19.18\% & 18.93\% & 6.56\\
FNO-2D \cite{FNO-ICLR-2021} & 1.28\% & 15.59\% & 15.56\% & 0.41\\
G.T. \cite{cao2021choose} & 0.94\% & \textbf{13.99\%} & \textbf{13.40\%} & 1.56 \\
OFormer \cite{li2023transformer} & 1.04\% & 17.55\% & 17.05\% & 1.85 \\
Ours & \textbf{0.88\%} & \textbf{10.43\%} & \textbf{12.48\%} & 1.02 \\ \hline
\end{tabular}
\label{table:results}
\end{table}

The NS1 dataset has a low Reynolds number and a long time horizon, making the primary challenge the error accumulation over time rather than complex and chaotic dynamics in the immediate future. Our model achieves an average test error of $0.88\%$ using multi-scale time-stepping with 4 dynamical models and a training rollout of 2 for each. Compared to a full rollout, the parallelization of training across time can be almost fully maintained, making it possible to speed up the training up to 10 times faster, since we also train four models. Moreover, the cost of the forward pass to predict the state at $t=49$ for example can be reduced up to 8 times since only 5 forward passes of $D^8$ are needed instead of 40 forward passes. The lesson here is that when the dynamics are simple enough, modeling large time-steps and a smaller training rollout can be a suitable strategy to achieve high accuracy with shorter and easier training.
\begin{figure}
\centering
\includegraphics[width=1\linewidth]{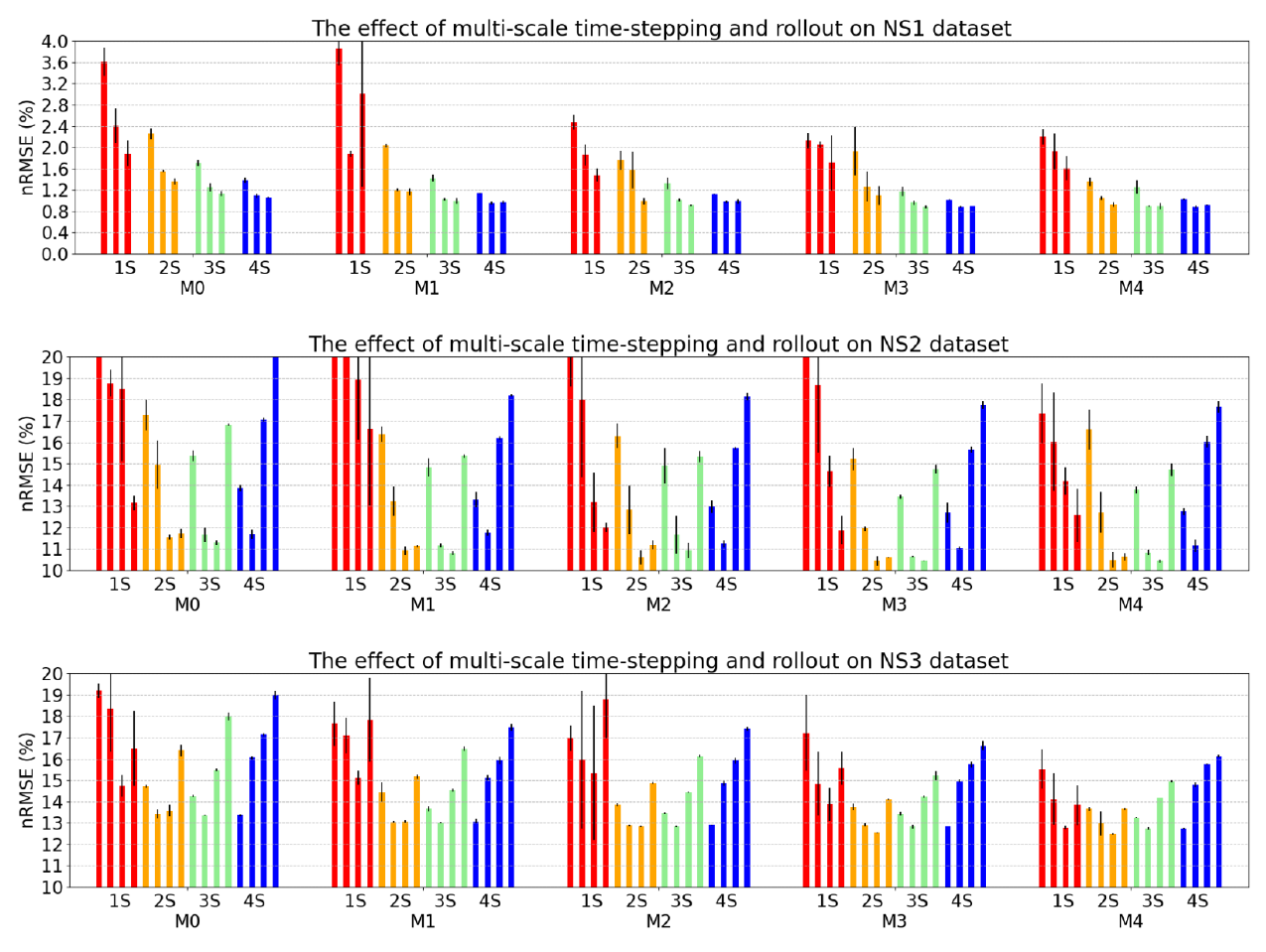}
  \caption{Multi-scale time-stepping and rollout effect on NS datasets. Each bar group represents a different type of attention mechanism from table \ref{table:hypers} denoted by M. Models of the same color use the same number of time scales denoted by their label (1S, 2S, 3S, 4S), in which different rollouts of $R=1,2,4,8$ are shown from left to right (except for $R=8$ excluded from NS1). The black error bars represent the variation across three random seeds that each model was trained with.}
\label{fig:ns_hp}
\end{figure}

The situation is different for NS2 and NS3 since they have more complex dynamics and shorter time horizons. The primary contributor to the prediction error here is the complex dynamics even in the immediate future. This is the reason FNO-3d, which outputs all future time-steps in a single forward pass, does not outperform FNO-2d as it does on NS1. The great performance of our model on these datasets implicates the remarkable potential of the architecture of its architecture and the enhancements introduced in this work to model complex spatiotemporal dynamics while having the Markov property.
\begin{table}
\centering
\caption{Explored options for attention and positional encoder}
\begin{tabular}{|c|c|c|} \hline
Name & Positional Encoding & Insertion Method \\ \hline
M0 & None & None \\ \hline
M1 & FFN(cartesian grid) & additive to input \\ \hline
M2 & FFN(periodic features) & additive to input \\ \hline
M3 & FFN(periodic features) & additive before softmax \\ \hline
M4 & FFN(periodic features) & multiplicative after softmax \\ \hline
\end{tabular}
\label{table:hypers}
\end{table}

In order to find the best choice for the number of time scales and the training rollout, as well as to verify the improvement of the model due to our proposed modifications, we experiment with 5 different types of attention mechanisms (Table \ref{table:hypers}) with different settings for multi-scale time-stepping and training rollout. Each model is trained with three different random seeds, and the average and standard deviation of the test error for each configuration are shown in Figure \ref{fig:ns_hp}. It is observed that M3 and M4 outperform the other options as expected because of the way they incorporate positional information without disturbing the actual input values for downstream calculations. The importance of including positional information and periodic positional features are also noticeable by looking at M0, M1, and M2.

Looking at the effect of the training rollout, it can be observed that low $R$ causes high uncertainty and variation based on the random seed. This means that even a finite rollout can be of great help to guide the training towards the global optima rather than a random local optima with poor performance when used autoregressively. However, if $R$ is too high considering the complexity of the system, the model may not succeed in simultaneously modeling the immediate future and a more distant future well, leading to an increase in the error for larger rollouts in NS2 and NS3. By finding the right balance, the responsibility of learning the dynamics of different time scales is divided across different models and lets them focus on different tasks. 

\begin{figure}
\centering
\includegraphics[width=1\linewidth]{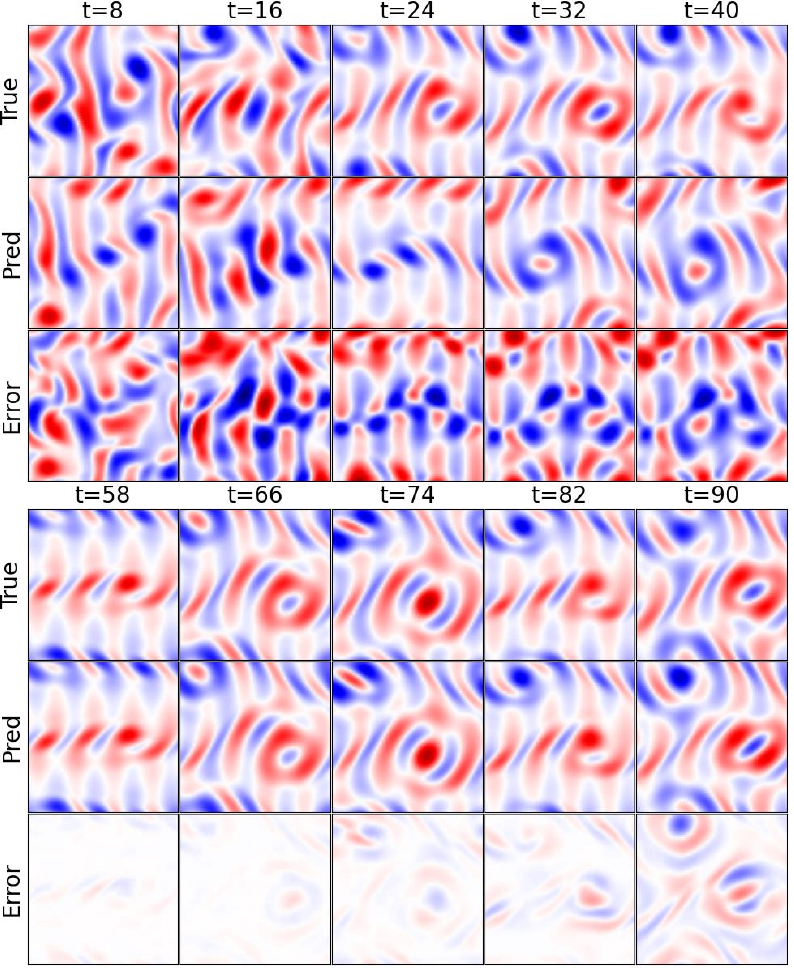}
  \caption{The model overfitting to the slow-changing regime in the Kolmogorov Flow and not learning the fast transient dynamics. Top: A rollout of $D^1$ for 40 time-steps, starting from $t_0=0$ (fast change). Bottom: A rollout of $D^1$ for 40 time-steps, starting from $t_0=50$ (slow change).}
\label{fig:KF}
\end{figure}

Moving on from the NS dataset, the results for the Kolmogorov Flow were not as successful but led to interesting observations. Although the final loss for both the autoencoder and the dynamical models were similar to NS3, our model failed to learn rapid changes accurately. However, it seems to perform well in the slow-changing regimes. Upon close investigation, this turned out to be because of the data distribution. Since the training is done across time, starting from arbitrary time-steps and predicting a finite number of future steps, it overfits the majority of the data which consists of a slow-changing and almost steady flow. However, the model does not show signs of instability even after 500 time-steps. This can indicate the possibility of learning models less prone to instability using the transformer architecture.

\section{Conclusion}

This work illustrated the potential of the transformer architecture to develop models to solve time-dependent PDEs that are efficient, accurate, and have the Markov property. By introducing effective enhancements to the transformer architecture and the training process, similar or better performance was observed compared to powerful models such as the Fourier Neural Operator or other transformer-based neural operators. By using a finite-horizon rollout in training, multi-scale time-stepping, and a new way to include positional information into the attention mechanism, our model was able to model the solution of 2D Navier-Stokes equations in turbulent regimes with a small training dataset, and fast training and testing. However, the model failed to learn the rapid dynamics in the Kolmogorov dataset and overfitted to the slow-changing dynamics.

In future work, coming up with a systematic and quick way of finding the best choice of training rollout or the number of time scales can be a promising venue for subsequent studies. Moreover, a theoretical investigation of the model and its weights and the connection between the dynamical models of different time scales may be able to provide new intuition and insight into the attention mechanism and its capabilities in modeling dynamic systems. Moreover, the modifications introduced in this work can be utilized on any model trying to learn the solution of time-dependent PDEs, as well as models based on the transformer architecture. We hope this work encourages researchers to move toward finding the best ways of incorporating the amazing transformer architecture in scientific computation applications.

\section{Acknowledgement}
This work is supported by the National Science Foundation under Grant No. 1953222.





\bibliographystyle{elsarticle-num-names} 
\bibliography{bibliography}

\begin{thebibliography}{65}
\expandafter\ifx\csname natexlab\endcsname\relax\def\natexlab#1{#1}\fi
\providecommand{\url}[1]{\texttt{#1}}
\providecommand{\href}[2]{#2}
\providecommand{\path}[1]{#1}
\providecommand{\DOIprefix}{doi:}
\providecommand{\ArXivprefix}{arXiv:}
\providecommand{\URLprefix}{URL: }
\providecommand{\Pubmedprefix}{pmid:}
\providecommand{\doi}[1]{\href{http://dx.doi.org/#1}{\path{#1}}}
\providecommand{\Pubmed}[1]{\href{pmid:#1}{\path{#1}}}
\providecommand{\bibinfo}[2]{#2}
\ifx\xfnm\relax \def\xfnm[#1]{\unskip,\space#1}\fi
\bibitem[{Brunton and Kutz(2019)}]{brunton2019data}
\bibinfo{author}{S.~L. Brunton}, \bibinfo{author}{J.~N. Kutz},
  \bibinfo{title}{Data-driven science and engineering: Machine learning,
  dynamical systems, and control}, \bibinfo{publisher}{Cambridge University
  Press}, \bibinfo{year}{2019}.
\bibitem[{Frank et~al.(2020)Frank, Drikakis, and Charissis}]{frank2020machine}
\bibinfo{author}{M.~Frank}, \bibinfo{author}{D.~Drikakis},
  \bibinfo{author}{V.~Charissis},
\newblock \bibinfo{title}{Machine-learning methods for computational science
  and engineering},
\newblock \bibinfo{journal}{Computation} \bibinfo{volume}{8}
  (\bibinfo{year}{2020}) \bibinfo{pages}{15}.
\bibitem[{Cuomo et~al.(2022)Cuomo, Di~Cola, Giampaolo, Rozza, Raissi, and
  Piccialli}]{cuomo2022scientific}
\bibinfo{author}{S.~Cuomo}, \bibinfo{author}{V.~S. Di~Cola},
  \bibinfo{author}{F.~Giampaolo}, \bibinfo{author}{G.~Rozza},
  \bibinfo{author}{M.~Raissi}, \bibinfo{author}{F.~Piccialli},
\newblock \bibinfo{title}{Scientific machine learning through physics--informed
  neural networks: Where we are and what’s next},
\newblock \bibinfo{journal}{Journal of Scientific Computing}
  \bibinfo{volume}{92} (\bibinfo{year}{2022}) \bibinfo{pages}{88}.
\bibitem[{Bhatnagar et~al.(2019)Bhatnagar, Afshar, Pan, Duraisamy, and
  Kaushik}]{Bhatnagar-CNN-PDE-2019}
\bibinfo{author}{S.~Bhatnagar}, \bibinfo{author}{Y.~Afshar},
  \bibinfo{author}{S.~Pan}, \bibinfo{author}{K.~Duraisamy},
  \bibinfo{author}{S.~Kaushik},
\newblock \bibinfo{title}{Prediction of aerodynamic flow fields using
  convolutional neural networks},
\newblock \bibinfo{journal}{Computational Mechanics} \bibinfo{volume}{64}
  (\bibinfo{year}{2019}) \bibinfo{pages}{525--545}. \URLprefix
  \url{https://doi.org/10.1007/s00466-019-01740-0}.
  \DOIprefix\doi{10.1007/s00466-019-01740-0}.
\bibitem[{Lee and Carlberg(2020)}]{lee2020model}
\bibinfo{author}{K.~Lee}, \bibinfo{author}{K.~T. Carlberg},
\newblock \bibinfo{title}{Model reduction of dynamical systems on nonlinear
  manifolds using deep convolutional autoencoders},
\newblock \bibinfo{journal}{Journal of Computational Physics}
  \bibinfo{volume}{404} (\bibinfo{year}{2020}) \bibinfo{pages}{108973}.
\bibitem[{Fukami et~al.(2020)Fukami, Nakamura, and
  Fukagata}]{fukami2020convolutional}
\bibinfo{author}{K.~Fukami}, \bibinfo{author}{T.~Nakamura},
  \bibinfo{author}{K.~Fukagata},
\newblock \bibinfo{title}{Convolutional neural network based hierarchical
  autoencoder for nonlinear mode decomposition of fluid field data},
\newblock \bibinfo{journal}{Physics of Fluids} \bibinfo{volume}{32}
  (\bibinfo{year}{2020}) \bibinfo{pages}{095110}.
\bibitem[{Gao et~al.(2021)Gao, Sun, and Wang}]{gao2021phygeonet}
\bibinfo{author}{H.~Gao}, \bibinfo{author}{L.~Sun}, \bibinfo{author}{J.-X.
  Wang},
\newblock \bibinfo{title}{Phygeonet: Physics-informed geometry-adaptive
  convolutional neural networks for solving parameterized steady-state pdes on
  irregular domain},
\newblock \bibinfo{journal}{Journal of Computational Physics}
  \bibinfo{volume}{428} (\bibinfo{year}{2021}) \bibinfo{pages}{110079}.
\bibitem[{Mohan and Gaitonde(2018)}]{mohan2018deep}
\bibinfo{author}{A.~T. Mohan}, \bibinfo{author}{D.~V. Gaitonde},
\newblock \bibinfo{title}{A deep learning based approach to reduced order
  modeling for turbulent flow control using lstm neural networks},
\newblock \bibinfo{journal}{arXiv preprint arXiv:1804.09269}
  (\bibinfo{year}{2018}).
\bibitem[{Hasegawa et~al.(2020)Hasegawa, Fukami, Murata, and
  Fukagata}]{hasegawa2020cnn}
\bibinfo{author}{K.~Hasegawa}, \bibinfo{author}{K.~Fukami},
  \bibinfo{author}{T.~Murata}, \bibinfo{author}{K.~Fukagata},
\newblock \bibinfo{title}{Cnn-lstm based reduced order modeling of
  two-dimensional unsteady flows around a circular cylinder at different
  reynolds numbers},
\newblock \bibinfo{journal}{Fluid Dynamics Research} \bibinfo{volume}{52}
  (\bibinfo{year}{2020}) \bibinfo{pages}{065501}.
\bibitem[{Maulik et~al.(2021)Maulik, Lusch, and
  Balaprakash}]{maulik2021reduced}
\bibinfo{author}{R.~Maulik}, \bibinfo{author}{B.~Lusch},
  \bibinfo{author}{P.~Balaprakash},
\newblock \bibinfo{title}{Reduced-order modeling of advection-dominated systems
  with recurrent neural networks and convolutional autoencoders},
\newblock \bibinfo{journal}{Physics of Fluids} \bibinfo{volume}{33}
  (\bibinfo{year}{2021}) \bibinfo{pages}{037106}.
\bibitem[{Vaswani et~al.(2017)Vaswani, Shazeer, Parmar, Uszkoreit, Jones,
  Gomez, Kaiser, and Polosukhin}]{Attention-NIPS-2017}
\bibinfo{author}{A.~Vaswani}, \bibinfo{author}{N.~Shazeer},
  \bibinfo{author}{N.~Parmar}, \bibinfo{author}{J.~Uszkoreit},
  \bibinfo{author}{L.~Jones}, \bibinfo{author}{A.~N. Gomez},
  \bibinfo{author}{L.~u. Kaiser}, \bibinfo{author}{I.~Polosukhin},
\newblock \bibinfo{title}{Attention is all you need},
\newblock in: \bibinfo{editor}{I.~Guyon}, \bibinfo{editor}{U.~V. Luxburg},
  \bibinfo{editor}{S.~Bengio}, \bibinfo{editor}{H.~Wallach},
  \bibinfo{editor}{R.~Fergus}, \bibinfo{editor}{S.~Vishwanathan},
  \bibinfo{editor}{R.~Garnett} (Eds.), \bibinfo{booktitle}{Advances in Neural
  Information Processing Systems}, volume~\bibinfo{volume}{30},
  \bibinfo{publisher}{Curran Associates, Inc.}, \bibinfo{year}{2017}.
\bibitem[{Devlin et~al.(2018)Devlin, Chang, Lee, and Toutanova}]{BERT}
\bibinfo{author}{J.~Devlin}, \bibinfo{author}{M.-W. Chang},
  \bibinfo{author}{K.~Lee}, \bibinfo{author}{K.~Toutanova},
  \bibinfo{title}{Bert: Pre-training of deep bidirectional transformers for
  language understanding}, \bibinfo{year}{2018}. \URLprefix
  \url{https://arxiv.org/abs/1810.04805}.
  \DOIprefix\doi{10.48550/ARXIV.1810.04805}.
\bibitem[{Tunyasuvunakool et~al.(2021)Tunyasuvunakool, Adler, Wu, Green,
  Zielinski, Žídek, Bridgland, Cowie, Meyer, Laydon, Velankar, Kleywegt,
  Bateman, Evans, Pritzel, Figurnov, Ronneberger, Bates, Kohl, and
  Hassabis}]{AlphaFold2}
\bibinfo{author}{K.~Tunyasuvunakool}, \bibinfo{author}{J.~Adler},
  \bibinfo{author}{Z.~Wu}, \bibinfo{author}{T.~Green},
  \bibinfo{author}{M.~Zielinski}, \bibinfo{author}{A.~Žídek},
  \bibinfo{author}{A.~Bridgland}, \bibinfo{author}{A.~Cowie},
  \bibinfo{author}{C.~Meyer}, \bibinfo{author}{A.~Laydon},
  \bibinfo{author}{S.~Velankar}, \bibinfo{author}{G.~Kleywegt},
  \bibinfo{author}{A.~Bateman}, \bibinfo{author}{R.~Evans},
  \bibinfo{author}{A.~Pritzel}, \bibinfo{author}{M.~Figurnov},
  \bibinfo{author}{O.~Ronneberger}, \bibinfo{author}{R.~Bates},
  \bibinfo{author}{S.~Kohl}, \bibinfo{author}{D.~Hassabis},
\newblock \bibinfo{title}{Highly accurate protein structure prediction for the
  human proteome},
\newblock \bibinfo{journal}{Nature} \bibinfo{volume}{596}
  (\bibinfo{year}{2021}) \bibinfo{pages}{1--9}.
  \DOIprefix\doi{10.1038/s41586-021-03828-1}.
\bibitem[{Dosovitskiy et~al.(2021)Dosovitskiy, Beyer, Kolesnikov, Weissenborn,
  Zhai, Unterthiner, Dehghani, Minderer, Heigold, Gelly, Uszkoreit, and
  Houlsby}]{Visual-Transformer-ICLR-2021}
\bibinfo{author}{A.~Dosovitskiy}, \bibinfo{author}{L.~Beyer},
  \bibinfo{author}{A.~Kolesnikov}, \bibinfo{author}{D.~Weissenborn},
  \bibinfo{author}{X.~Zhai}, \bibinfo{author}{T.~Unterthiner},
  \bibinfo{author}{M.~Dehghani}, \bibinfo{author}{M.~Minderer},
  \bibinfo{author}{G.~Heigold}, \bibinfo{author}{S.~Gelly},
  \bibinfo{author}{J.~Uszkoreit}, \bibinfo{author}{N.~Houlsby},
\newblock \bibinfo{title}{An image is worth 16x16 words: Transformers for image
  recognition at scale},
\newblock in: \bibinfo{booktitle}{International Conference on Learning
  Representations}, \bibinfo{year}{2021}. \URLprefix
  \url{https://openreview.net/forum?id=YicbFdNTTy}.
\bibitem[{Berkooz et~al.(1993)Berkooz, Holmes, and Lumley}]{berkooz1993proper}
\bibinfo{author}{G.~Berkooz}, \bibinfo{author}{P.~Holmes},
  \bibinfo{author}{J.~L. Lumley},
\newblock \bibinfo{title}{The proper orthogonal decomposition in the analysis
  of turbulent flows},
\newblock \bibinfo{journal}{Annual review of fluid mechanics}
  \bibinfo{volume}{25} (\bibinfo{year}{1993}) \bibinfo{pages}{539--575}.
\bibitem[{Anttonen et~al.(2003)Anttonen, King, and Beran}]{anttonen2003pod}
\bibinfo{author}{J.~S. Anttonen}, \bibinfo{author}{P.~I. King},
  \bibinfo{author}{P.~S. Beran},
\newblock \bibinfo{title}{Pod-based reduced-order models with deforming grids},
\newblock \bibinfo{journal}{Mathematical and Computer Modelling}
  \bibinfo{volume}{38} (\bibinfo{year}{2003}) \bibinfo{pages}{41--62}.
\bibitem[{Murata et~al.(2020)Murata, Fukami, and
  Fukagata}]{murata2020nonlinear}
\bibinfo{author}{T.~Murata}, \bibinfo{author}{K.~Fukami},
  \bibinfo{author}{K.~Fukagata},
\newblock \bibinfo{title}{Nonlinear mode decomposition with convolutional
  neural networks for fluid dynamics},
\newblock \bibinfo{journal}{Journal of Fluid Mechanics} \bibinfo{volume}{882}
  (\bibinfo{year}{2020}) \bibinfo{pages}{A13}.
\bibitem[{Hesthaven and Ubbiali(2018)}]{hesthaven2018non}
\bibinfo{author}{J.~S. Hesthaven}, \bibinfo{author}{S.~Ubbiali},
\newblock \bibinfo{title}{Non-intrusive reduced order modeling of nonlinear
  problems using neural networks},
\newblock \bibinfo{journal}{Journal of Computational Physics}
  \bibinfo{volume}{363} (\bibinfo{year}{2018}) \bibinfo{pages}{55--78}.
\bibitem[{Brunton et~al.(2020)Brunton, Noack, and
  Koumoutsakos}]{brunton2020machine}
\bibinfo{author}{S.~L. Brunton}, \bibinfo{author}{B.~R. Noack},
  \bibinfo{author}{P.~Koumoutsakos},
\newblock \bibinfo{title}{Machine learning for fluid mechanics},
\newblock \bibinfo{journal}{Annual review of fluid mechanics}
  \bibinfo{volume}{52} (\bibinfo{year}{2020}) \bibinfo{pages}{477--508}.
\bibitem[{Wiewel et~al.(2019)Wiewel, Becher, and Thuerey}]{wiewel2019latent}
\bibinfo{author}{S.~Wiewel}, \bibinfo{author}{M.~Becher},
  \bibinfo{author}{N.~Thuerey},
\newblock \bibinfo{title}{Latent space physics: Towards learning the temporal
  evolution of fluid flow},
\newblock in: \bibinfo{booktitle}{Computer graphics forum},
  volume~\bibinfo{volume}{38}, \bibinfo{organization}{Wiley Online Library},
  \bibinfo{year}{2019}, pp. \bibinfo{pages}{71--82}.
\bibitem[{Lee and Carlberg(2021)}]{lee2021deep}
\bibinfo{author}{K.~Lee}, \bibinfo{author}{K.~T. Carlberg},
\newblock \bibinfo{title}{Deep conservation: A latent-dynamics model for exact
  satisfaction of physical conservation laws},
\newblock in: \bibinfo{booktitle}{Proceedings of the AAAI Conference on
  Artificial Intelligence}, volume~\bibinfo{volume}{35}, \bibinfo{year}{2021},
  pp. \bibinfo{pages}{277--285}.
\bibitem[{Khoo et~al.(2021)Khoo, Lu, and Ying}]{khoo2021solving}
\bibinfo{author}{Y.~Khoo}, \bibinfo{author}{J.~Lu}, \bibinfo{author}{L.~Ying},
\newblock \bibinfo{title}{Solving parametric pde problems with artificial
  neural networks},
\newblock \bibinfo{journal}{European Journal of Applied Mathematics}
  \bibinfo{volume}{32} (\bibinfo{year}{2021}) \bibinfo{pages}{421--435}.
\bibitem[{Gupta and Brandstetter(2022)}]{gupta2022multispatiotemporalscale}
\bibinfo{author}{J.~K. Gupta}, \bibinfo{author}{J.~Brandstetter},
  \bibinfo{title}{Towards multi-spatiotemporal-scale generalized pde modeling},
  \bibinfo{year}{2022}. \href{http://arxiv.org/abs/2209.15616}{{\tt
  arXiv:2209.15616}}.
\bibitem[{JANNY et~al.(2023)JANNY, B{\'e}n{\'e}teau, Nadri, Digne, THOME, and
  Wolf}]{janny2023eagle}
\bibinfo{author}{S.~JANNY}, \bibinfo{author}{A.~B{\'e}n{\'e}teau},
  \bibinfo{author}{M.~Nadri}, \bibinfo{author}{J.~Digne},
  \bibinfo{author}{N.~THOME}, \bibinfo{author}{C.~Wolf},
\newblock \bibinfo{title}{{EAGLE}: Large-scale learning of turbulent fluid
  dynamics with mesh transformers},
\newblock in: \bibinfo{booktitle}{The Eleventh International Conference on
  Learning Representations}, \bibinfo{year}{2023}. \URLprefix
  \url{https://openreview.net/forum?id=mfIX4QpsARJ}.
\bibitem[{Li and Farimani(2022)}]{li2022fgn}
\bibinfo{author}{Z.~Li}, \bibinfo{author}{A.~B. Farimani},
\newblock \bibinfo{title}{Graph neural network-accelerated lagrangian fluid
  simulation},
\newblock \bibinfo{journal}{Computers \& Graphics} \bibinfo{volume}{103}
  (\bibinfo{year}{2022}) \bibinfo{pages}{201--211}.
  \DOIprefix\doi{https://doi.org/10.1016/j.cag.2022.02.004}.
\bibitem[{Pant et~al.(2021)Pant, Doshi, Bahl, and
  Barati~Farimani}]{PDE-ROM-2021}
\bibinfo{author}{P.~Pant}, \bibinfo{author}{R.~Doshi},
  \bibinfo{author}{P.~Bahl}, \bibinfo{author}{A.~Barati~Farimani},
\newblock \bibinfo{title}{Deep learning for reduced order modelling and
  efficient temporal evolution of fluid simulations},
\newblock \bibinfo{journal}{Physics of Fluids} \bibinfo{volume}{33}
  (\bibinfo{year}{2021}) \bibinfo{pages}{107101}. \URLprefix
  \url{https://doi.org/10.1063/5.0062546}. \DOIprefix\doi{10.1063/5.0062546}.
  \href{http://arxiv.org/abs/https://doi.org/10.1063/5.0062546}{{\tt
  arXiv:https://doi.org/10.1063/5.0062546}}.
\bibitem[{Pfaff et~al.(2021)Pfaff, Fortunato, Sanchez-Gonzalez, and
  Battaglia}]{pfaff2021learning}
\bibinfo{author}{T.~Pfaff}, \bibinfo{author}{M.~Fortunato},
  \bibinfo{author}{A.~Sanchez-Gonzalez}, \bibinfo{author}{P.~W. Battaglia},
  \bibinfo{title}{Learning mesh-based simulation with graph networks},
  \bibinfo{year}{2021}. \href{http://arxiv.org/abs/2010.03409}{{\tt
  arXiv:2010.03409}}.
\bibitem[{Sanchez-Gonzalez et~al.(2020)Sanchez-Gonzalez, Godwin, Pfaff, Ying,
  Leskovec, and Battaglia}]{sanchezgonzalez2020learning}
\bibinfo{author}{A.~Sanchez-Gonzalez}, \bibinfo{author}{J.~Godwin},
  \bibinfo{author}{T.~Pfaff}, \bibinfo{author}{R.~Ying},
  \bibinfo{author}{J.~Leskovec}, \bibinfo{author}{P.~W. Battaglia},
  \bibinfo{title}{Learning to simulate complex physics with graph networks},
  \bibinfo{year}{2020}. \href{http://arxiv.org/abs/2002.09405}{{\tt
  arXiv:2002.09405}}.
\bibitem[{Belbute-Peres et~al.(2020)Belbute-Peres, Economon, and
  Kolter}]{belbute2020combining}
\bibinfo{author}{F.~D.~A. Belbute-Peres}, \bibinfo{author}{T.~Economon},
  \bibinfo{author}{Z.~Kolter},
\newblock \bibinfo{title}{Combining differentiable pde solvers and graph neural
  networks for fluid flow prediction},
\newblock in: \bibinfo{booktitle}{international conference on machine
  learning}, \bibinfo{organization}{PMLR}, \bibinfo{year}{2020}, pp.
  \bibinfo{pages}{2402--2411}.
\bibitem[{Raissi et~al.(2019)Raissi, Perdikaris, and
  Karniadakis}]{raissi2019physics}
\bibinfo{author}{M.~Raissi}, \bibinfo{author}{P.~Perdikaris},
  \bibinfo{author}{G.~E. Karniadakis},
\newblock \bibinfo{title}{Physics-informed neural networks: A deep learning
  framework for solving forward and inverse problems involving nonlinear
  partial differential equations},
\newblock \bibinfo{journal}{Journal of Computational physics}
  \bibinfo{volume}{378} (\bibinfo{year}{2019}) \bibinfo{pages}{686--707}.
\bibitem[{Karniadakis et~al.(2021)Karniadakis, Kevrekidis, Lu, Perdikaris,
  Wang, and Yang}]{physics-informed-ml-review}
\bibinfo{author}{G.~Karniadakis}, \bibinfo{author}{Y.~Kevrekidis},
  \bibinfo{author}{L.~Lu}, \bibinfo{author}{P.~Perdikaris},
  \bibinfo{author}{S.~Wang}, \bibinfo{author}{L.~Yang},
\newblock \bibinfo{title}{Physics-informed machine learning}
  (\bibinfo{year}{2021}) \bibinfo{pages}{1--19}.
  \DOIprefix\doi{10.1038/s42254-021-00314-5}.
\bibitem[{Cai et~al.(2021)Cai, Mao, Wang, Yin, and
  Karniadakis}]{cai2021physics}
\bibinfo{author}{S.~Cai}, \bibinfo{author}{Z.~Mao}, \bibinfo{author}{Z.~Wang},
  \bibinfo{author}{M.~Yin}, \bibinfo{author}{G.~E. Karniadakis},
\newblock \bibinfo{title}{Physics-informed neural networks (pinns) for fluid
  mechanics: A review},
\newblock \bibinfo{journal}{Acta Mechanica Sinica} \bibinfo{volume}{37}
  (\bibinfo{year}{2021}) \bibinfo{pages}{1727--1738}.
\bibitem[{Pang et~al.(2019)Pang, Lu, and Karniadakis}]{pang2019fpinns}
\bibinfo{author}{G.~Pang}, \bibinfo{author}{L.~Lu}, \bibinfo{author}{G.~E.
  Karniadakis},
\newblock \bibinfo{title}{fpinns: Fractional physics-informed neural networks},
\newblock \bibinfo{journal}{SIAM Journal on Scientific Computing}
  \bibinfo{volume}{41} (\bibinfo{year}{2019}) \bibinfo{pages}{A2603--A2626}.
\bibitem[{Sun et~al.(2020)Sun, Gao, Pan, and Wang}]{sun2020surrogate}
\bibinfo{author}{L.~Sun}, \bibinfo{author}{H.~Gao}, \bibinfo{author}{S.~Pan},
  \bibinfo{author}{J.-X. Wang},
\newblock \bibinfo{title}{Surrogate modeling for fluid flows based on
  physics-constrained deep learning without simulation data},
\newblock \bibinfo{journal}{Computer Methods in Applied Mechanics and
  Engineering} \bibinfo{volume}{361} (\bibinfo{year}{2020})
  \bibinfo{pages}{112732}.
\bibitem[{Chen and Chen(1995)}]{Universal-apprx-operator-IEEE-1995}
\bibinfo{author}{T.~Chen}, \bibinfo{author}{H.~Chen},
\newblock \bibinfo{title}{Universal approximation to nonlinear operators by
  neural networks with arbitrary activation functions and its application to
  dynamical systems},
\newblock \bibinfo{journal}{IEEE Transactions on Neural Networks}
  \bibinfo{volume}{6} (\bibinfo{year}{1995}) \bibinfo{pages}{911--917}.
  \DOIprefix\doi{10.1109/72.392253}.
\bibitem[{Lu et~al.(2021)Lu, Jin, Pang, Zhang, and
  Karniadakis}]{DeepONet-Nature-2021}
\bibinfo{author}{L.~Lu}, \bibinfo{author}{P.~Jin}, \bibinfo{author}{G.~Pang},
  \bibinfo{author}{Z.~Zhang}, \bibinfo{author}{G.~E. Karniadakis},
\newblock \bibinfo{title}{Learning nonlinear operators via deeponet based on
  the universal approximation theorem of operators},
\newblock \bibinfo{journal}{Nature Machine Intelligence} \bibinfo{volume}{3}
  (\bibinfo{year}{2021}) \bibinfo{pages}{218--229}. \URLprefix
  \url{https://doi.org/10.1038/s42256-021-00302-5}.
  \DOIprefix\doi{10.1038/s42256-021-00302-5}.
\bibitem[{Li et~al.(2020)Li, Kovachki, Azizzadenesheli, Liu, Bhattacharya,
  Stuart, and Anandkumar}]{li2020gno}
\bibinfo{author}{Z.~Li}, \bibinfo{author}{N.~Kovachki},
  \bibinfo{author}{K.~Azizzadenesheli}, \bibinfo{author}{B.~Liu},
  \bibinfo{author}{K.~Bhattacharya}, \bibinfo{author}{A.~Stuart},
  \bibinfo{author}{A.~Anandkumar}, \bibinfo{title}{Neural operator: Graph
  kernel network for partial differential equations}, \bibinfo{year}{2020}.
  \href{http://arxiv.org/abs/2003.03485}{{\tt arXiv:2003.03485}}.
\bibitem[{Cao(2021)}]{cao2021choose}
\bibinfo{author}{S.~Cao},
\newblock \bibinfo{title}{Choose a transformer: Fourier or galerkin},
\newblock \bibinfo{journal}{Advances in neural information processing systems}
  \bibinfo{volume}{34} (\bibinfo{year}{2021}) \bibinfo{pages}{24924--24940}.
\bibitem[{Li et~al.(2022)Li, Meidani, and Farimani}]{li2022transformer}
\bibinfo{author}{Z.~Li}, \bibinfo{author}{K.~Meidani}, \bibinfo{author}{A.~B.
  Farimani},
\newblock \bibinfo{title}{Transformer for partial differential equations'
  operator learning},
\newblock \bibinfo{journal}{arXiv preprint arXiv:2205.13671}
  (\bibinfo{year}{2022}).
\bibitem[{Gupta et~al.(2021)Gupta, Xiao, and
  Bogdan}]{gupta2021multiwaveletbased}
\bibinfo{author}{G.~Gupta}, \bibinfo{author}{X.~Xiao},
  \bibinfo{author}{P.~Bogdan}, \bibinfo{title}{Multiwavelet-based operator
  learning for differential equations}, \bibinfo{year}{2021}.
  \href{http://arxiv.org/abs/2109.13459}{{\tt arXiv:2109.13459}}.
\bibitem[{Li et~al.(2020)Li, Kovachki, Azizzadenesheli, Liu, Bhattacharya,
  Stuart, and Anandkumar}]{li2020fourier}
\bibinfo{author}{Z.~Li}, \bibinfo{author}{N.~Kovachki},
  \bibinfo{author}{K.~Azizzadenesheli}, \bibinfo{author}{B.~Liu},
  \bibinfo{author}{K.~Bhattacharya}, \bibinfo{author}{A.~Stuart},
  \bibinfo{author}{A.~Anandkumar},
\newblock \bibinfo{title}{Fourier neural operator for parametric partial
  differential equations},
\newblock \bibinfo{journal}{arXiv preprint arXiv:2010.08895}
  (\bibinfo{year}{2020}).
\bibitem[{Wen et~al.(2022)Wen, Li, Azizzadenesheli, Anandkumar, and
  Benson}]{wen2022ufno}
\bibinfo{author}{G.~Wen}, \bibinfo{author}{Z.~Li},
  \bibinfo{author}{K.~Azizzadenesheli}, \bibinfo{author}{A.~Anandkumar},
  \bibinfo{author}{S.~M. Benson},
\newblock \bibinfo{title}{U-fno—an enhanced fourier neural operator-based
  deep-learning model for multiphase flow},
\newblock \bibinfo{journal}{Advances in Water Resources} \bibinfo{volume}{163}
  (\bibinfo{year}{2022}) \bibinfo{pages}{104180}.
\bibitem[{Tran et~al.(2023)Tran, Mathews, Xie, and Ong}]{tran2023factfno}
\bibinfo{author}{A.~Tran}, \bibinfo{author}{A.~Mathews},
  \bibinfo{author}{L.~Xie}, \bibinfo{author}{C.~S. Ong},
  \bibinfo{title}{Factorized fourier neural operators}, \bibinfo{year}{2023}.
  \href{http://arxiv.org/abs/2111.13802}{{\tt arXiv:2111.13802}}.
\bibitem[{Carion et~al.(2020)Carion, Massa, Synnaeve, Usunier, Kirillov, and
  Zagoruyko}]{carion2020detr}
\bibinfo{author}{N.~Carion}, \bibinfo{author}{F.~Massa},
  \bibinfo{author}{G.~Synnaeve}, \bibinfo{author}{N.~Usunier},
  \bibinfo{author}{A.~Kirillov}, \bibinfo{author}{S.~Zagoruyko},
  \bibinfo{title}{End-to-end object detection with transformers},
  \bibinfo{year}{2020}. \href{http://arxiv.org/abs/2005.12872}{{\tt
  arXiv:2005.12872}}.
\bibitem[{Geneva and Zabaras(2022)}]{geneva2022transformers}
\bibinfo{author}{N.~Geneva}, \bibinfo{author}{N.~Zabaras},
\newblock \bibinfo{title}{Transformers for modeling physical systems},
\newblock \bibinfo{journal}{Neural Networks} \bibinfo{volume}{146}
  (\bibinfo{year}{2022}) \bibinfo{pages}{272--289}.
\bibitem[{Solera-Rico et~al.(2023)Solera-Rico, Vila, G\'omez, Wang, Almashjary,
  Dawson, and Vinuesa}]{solera2023beta}
\bibinfo{author}{A.~Solera-Rico}, \bibinfo{author}{C.~S. Vila},
  \bibinfo{author}{M.~G\'omez}, \bibinfo{author}{Y.~Wang},
  \bibinfo{author}{A.~Almashjary}, \bibinfo{author}{S.~Dawson},
  \bibinfo{author}{R.~Vinuesa},
\newblock \bibinfo{title}{$\beta$-variational autoencoders and transformers for
  reduced-order modeling of fluid flows},
\newblock \bibinfo{journal}{arXiv preprint arXiv:2304.03571}
  (\bibinfo{year}{2023}).
\bibitem[{Hemmasian and Barati~Farimani(2023)}]{hemmasian2023reduced}
\bibinfo{author}{A.~Hemmasian}, \bibinfo{author}{A.~Barati~Farimani},
\newblock \bibinfo{title}{Reduced-order modeling of fluid flows with
  transformers},
\newblock \bibinfo{journal}{Physics of Fluids} \bibinfo{volume}{35}
  (\bibinfo{year}{2023}).
\bibitem[{Guo et~al.(2022)Guo, Cao, and Chen}]{guo2022transformer}
\bibinfo{author}{R.~Guo}, \bibinfo{author}{S.~Cao}, \bibinfo{author}{L.~Chen},
\newblock \bibinfo{title}{Transformer meets boundary value inverse problems},
\newblock \bibinfo{journal}{arXiv preprint arXiv:2209.14977}
  (\bibinfo{year}{2022}).
\bibitem[{Hao et~al.(2023)Hao, Ying, Wang, Su, Dong, Liu, Cheng, Zhu, and
  Song}]{hao2023gnot}
\bibinfo{author}{Z.~Hao}, \bibinfo{author}{C.~Ying}, \bibinfo{author}{Z.~Wang},
  \bibinfo{author}{H.~Su}, \bibinfo{author}{Y.~Dong}, \bibinfo{author}{S.~Liu},
  \bibinfo{author}{Z.~Cheng}, \bibinfo{author}{J.~Zhu},
  \bibinfo{author}{J.~Song},
\newblock \bibinfo{title}{Gnot: A general neural operator transformer for
  operator learning},
\newblock \bibinfo{journal}{arXiv preprint arXiv:2302.14376}
  (\bibinfo{year}{2023}).
\bibitem[{Li et~al.(2023)Li, Meidani, and Farimani}]{li2023transformer}
\bibinfo{author}{Z.~Li}, \bibinfo{author}{K.~Meidani}, \bibinfo{author}{A.~B.
  Farimani},
\newblock \bibinfo{title}{Transformer for partial differential equations
  operator learning},
\newblock \bibinfo{journal}{Transactions on Machine Learning Research}
  (\bibinfo{year}{2023}). \URLprefix
  \url{https://openreview.net/forum?id=EPPqt3uERT}.
\bibitem[{Ovadia et~al.(2023)Ovadia, Kahana, Stinis, Turkel, and
  Karniadakis}]{ovadia2023vito}
\bibinfo{author}{O.~Ovadia}, \bibinfo{author}{A.~Kahana},
  \bibinfo{author}{P.~Stinis}, \bibinfo{author}{E.~Turkel},
  \bibinfo{author}{G.~E. Karniadakis},
\newblock \bibinfo{title}{Vito: Vision transformer-operator},
\newblock \bibinfo{journal}{arXiv preprint arXiv:2303.08891}
  (\bibinfo{year}{2023}).
\bibitem[{Kissas et~al.(2022)Kissas, Seidman, Guilhoto, Preciado, Pappas, and
  Perdikaris}]{kissas2022learning}
\bibinfo{author}{G.~Kissas}, \bibinfo{author}{J.~Seidman},
  \bibinfo{author}{L.~F. Guilhoto}, \bibinfo{author}{V.~M. Preciado},
  \bibinfo{author}{G.~J. Pappas}, \bibinfo{author}{P.~Perdikaris},
  \bibinfo{title}{Learning operators with coupled attention},
  \bibinfo{year}{2022}. \href{http://arxiv.org/abs/2201.01032}{{\tt
  arXiv:2201.01032}}.
\bibitem[{Liu et~al.(2022)Liu, Xu, and Zhang}]{liu2022ht}
\bibinfo{author}{X.~Liu}, \bibinfo{author}{B.~Xu}, \bibinfo{author}{L.~Zhang},
\newblock \bibinfo{title}{Ht-net: Hierarchical transformer based operator
  learning model for multiscale pdes},
\newblock \bibinfo{journal}{arXiv preprint arXiv:2210.10890}
  (\bibinfo{year}{2022}).
\bibitem[{Nguyen et~al.(2022)Nguyen, Pham, Nguyen, Nguyen, Osher, and
  Ho}]{nguyen2022fourierformer}
\bibinfo{author}{T.~Nguyen}, \bibinfo{author}{M.~Pham},
  \bibinfo{author}{T.~Nguyen}, \bibinfo{author}{K.~Nguyen},
  \bibinfo{author}{S.~Osher}, \bibinfo{author}{N.~Ho},
\newblock \bibinfo{title}{Fourierformer: Transformer meets generalized fourier
  integral theorem},
\newblock \bibinfo{journal}{Advances in Neural Information Processing Systems}
  \bibinfo{volume}{35} (\bibinfo{year}{2022}) \bibinfo{pages}{29319--29335}.
\bibitem[{Li et~al.(2023)Li, Shu, and Farimani}]{li2023scalable}
\bibinfo{author}{Z.~Li}, \bibinfo{author}{D.~Shu}, \bibinfo{author}{A.~B.
  Farimani},
\newblock \bibinfo{title}{Scalable transformer for pde surrogate modeling},
\newblock \bibinfo{journal}{arXiv preprint arXiv:2305.17560}
  (\bibinfo{year}{2023}).
\bibitem[{Han et~al.(2022)Han, Gao, Pfaff, Wang, and Liu}]{han2022predicting}
\bibinfo{author}{X.~Han}, \bibinfo{author}{H.~Gao}, \bibinfo{author}{T.~Pfaff},
  \bibinfo{author}{J.-X. Wang}, \bibinfo{author}{L.-P. Liu},
\newblock \bibinfo{title}{Predicting physics in mesh-reduced space with
  temporal attention},
\newblock \bibinfo{journal}{arXiv preprint arXiv:2201.09113}
  (\bibinfo{year}{2022}).
\bibitem[{Rae et~al.(2019)Rae, Potapenko, Jayakumar, and
  Lillicrap}]{rae2019compressive}
\bibinfo{author}{J.~W. Rae}, \bibinfo{author}{A.~Potapenko},
  \bibinfo{author}{S.~M. Jayakumar}, \bibinfo{author}{T.~P. Lillicrap},
\newblock \bibinfo{title}{Compressive transformers for long-range sequence
  modelling},
\newblock \bibinfo{journal}{arXiv preprint arXiv:1911.05507}
  (\bibinfo{year}{2019}).
\bibitem[{Zaheer et~al.(2021)Zaheer, Guruganesh, Dubey, Ainslie, Alberti,
  Ontanon, Pham, Ravula, Wang, Yang, and Ahmed}]{zaheer2021bigbird}
\bibinfo{author}{M.~Zaheer}, \bibinfo{author}{G.~Guruganesh},
  \bibinfo{author}{A.~Dubey}, \bibinfo{author}{J.~Ainslie},
  \bibinfo{author}{C.~Alberti}, \bibinfo{author}{S.~Ontanon},
  \bibinfo{author}{P.~Pham}, \bibinfo{author}{A.~Ravula},
  \bibinfo{author}{Q.~Wang}, \bibinfo{author}{L.~Yang},
  \bibinfo{author}{A.~Ahmed}, \bibinfo{title}{Big bird: Transformers for longer
  sequences}, \bibinfo{year}{2021}. \href{http://arxiv.org/abs/2007.14062}{{\tt
  arXiv:2007.14062}}.
\bibitem[{Beltagy et~al.(2020)Beltagy, Peters, and
  Cohan}]{beltagy2020longformer}
\bibinfo{author}{I.~Beltagy}, \bibinfo{author}{M.~E. Peters},
  \bibinfo{author}{A.~Cohan}, \bibinfo{title}{Longformer: The long-document
  transformer}, \bibinfo{year}{2020}.
  \href{http://arxiv.org/abs/2004.05150}{{\tt arXiv:2004.05150}}.
\bibitem[{Kitaev et~al.(2020)Kitaev, Kaiser, and Levskaya}]{Kitaev2020Reformer}
\bibinfo{author}{N.~Kitaev}, \bibinfo{author}{L.~Kaiser},
  \bibinfo{author}{A.~Levskaya},
\newblock \bibinfo{title}{Reformer: The efficient transformer},
\newblock in: \bibinfo{booktitle}{International Conference on Learning
  Representations}, \bibinfo{year}{2020}. \URLprefix
  \url{https://openreview.net/forum?id=rkgNKkHtvB}.
\bibitem[{Shen et~al.(2020)Shen, Zhang, Zhao, Yi, and Li}]{shen2020efficient}
\bibinfo{author}{Z.~Shen}, \bibinfo{author}{M.~Zhang},
  \bibinfo{author}{H.~Zhao}, \bibinfo{author}{S.~Yi}, \bibinfo{author}{H.~Li},
  \bibinfo{title}{Efficient attention: Attention with linear complexities},
  \bibinfo{year}{2020}. \href{http://arxiv.org/abs/1812.01243}{{\tt
  arXiv:1812.01243}}.
\bibitem[{Li et~al.(2021)Li, Kovachki, Azizzadenesheli, liu, Bhattacharya,
  Stuart, and Anandkumar}]{FNO-ICLR-2021}
\bibinfo{author}{Z.~Li}, \bibinfo{author}{N.~B. Kovachki},
  \bibinfo{author}{K.~Azizzadenesheli}, \bibinfo{author}{B.~liu},
  \bibinfo{author}{K.~Bhattacharya}, \bibinfo{author}{A.~Stuart},
  \bibinfo{author}{A.~Anandkumar},
\newblock \bibinfo{title}{Fourier neural operator for parametric partial
  differential equations},
\newblock in: \bibinfo{booktitle}{International Conference on Learning
  Representations}, \bibinfo{year}{2021}. \URLprefix
  \url{https://openreview.net/forum?id=c8P9NQVtmnO}.
\bibitem[{Liu et~al.(2022)Liu, Kutz, and Brunton}]{liu2022hierarchical}
\bibinfo{author}{Y.~Liu}, \bibinfo{author}{J.~N. Kutz}, \bibinfo{author}{S.~L.
  Brunton},
\newblock \bibinfo{title}{Hierarchical deep learning of multiscale differential
  equation time-steppers},
\newblock \bibinfo{journal}{Philosophical Transactions of the Royal Society A}
  \bibinfo{volume}{380} (\bibinfo{year}{2022}) \bibinfo{pages}{20210200}.
\bibitem[{Kingma and Ba(2014)}]{Adam}
\bibinfo{author}{D.~P. Kingma}, \bibinfo{author}{J.~Ba}, \bibinfo{title}{Adam:
  A method for stochastic optimization}, \bibinfo{year}{2014}. \URLprefix
  \url{https://arxiv.org/abs/1412.6980}.
  \DOIprefix\doi{10.48550/ARXIV.1412.6980}.
\bibitem[{Wang et~al.(2020)Wang, Kashinath, Mustafa, Albert, and
  Yu}]{TF-NET-SIGKDD-2020}
\bibinfo{author}{R.~Wang}, \bibinfo{author}{K.~Kashinath},
  \bibinfo{author}{M.~Mustafa}, \bibinfo{author}{A.~Albert},
  \bibinfo{author}{R.~Yu},
\newblock \bibinfo{title}{Towards physics-informed deep learning for turbulent
  flow prediction},
\newblock in: \bibinfo{booktitle}{Proceedings of the 26th ACM SIGKDD
  International Conference on Knowledge Discovery \& Data Mining}, KDD '20,
  \bibinfo{publisher}{Association for Computing Machinery},
  \bibinfo{address}{New York, NY, USA}, \bibinfo{year}{2020}, p.
  \bibinfo{pages}{1457–1466}. \URLprefix
  \url{https://doi.org/10.1145/3394486.3403198}.
  \DOIprefix\doi{10.1145/3394486.3403198}.

\end{thebibliography}


\end{document}